\documentclass{article}
\usepackage{ijcai21}

\usepackage{times}

\usepackage{soul}
\usepackage{url}
\usepackage[hidelinks]{hyperref}
\usepackage[utf8]{inputenc}
\usepackage[small]{caption}
\usepackage{graphicx}
\usepackage{amsmath}
\usepackage{booktabs}
\usepackage{algorithm}
\usepackage{algorithmic}
\usepackage{subfiles}
\usepackage{enumitem}
\usepackage{float}
\usepackage{booktabs}
\usepackage{multirow}
\usepackage{subcaption}
\usepackage{epsfig}
\usepackage{amssymb}

\title{Distilled Replay: Overcoming Forgetting through Synthetic Samples}

\author{
Andrea Rosasco$^1$\footnote{Contact Author}\and
Antonio Carta$^1$\and
Andrea Cossu$^2$\and
Vincenzo Lomonaco$^1$\And
Davide Bacciu$^1$\\
\affiliations
$^1$University of Pisa\\
$^2$Scuola Normale Superiore\\
}

\begin{document}

\maketitle

\begin{abstract}
Replay strategies are Continual Learning techniques which mitigate catastrophic forgetting by keeping a buffer of patterns from previous experiences, which are interleaved with new data during training. The amount of patterns stored in the buffer is a critical parameter which largely influences the final performance and the memory footprint of the approach. This work introduces Distilled Replay, a novel replay strategy for Continual Learning which is able to mitigate forgetting by keeping a very small buffer ($1$ pattern per class) of highly informative samples. Distilled Replay builds the buffer through a distillation process which compresses a large dataset into a tiny set of informative examples. We show the effectiveness of our Distilled Replay against popular replay-based strategies on four Continual Learning benchmarks.
\end{abstract}

\section{Introduction}
Deep learning models trained under the assumption that all training data is available from the beginning and each sample is independent and identically distributed manage to achieve impressive performance \cite{Krizhevsky2012}. This learning scenario is often called \emph{offline training}. Contrary to offline training, \emph{continual learning} (CL) requires the model to learn sequentially from a stream of experiences \cite{lesort2020}. Each experience is made of a batch of data, which may contain new knowledge such as novel classes that need to be distinguished by the model. Therefore, the model must be continually updated to incorporate knowledge coming from new experiences. However, when trained on new samples, neural networks tend to forget past knowledge: this phenomenon is called catastrophic forgetting \cite{french1999}. Catastrophic forgetting emerges as a consequence of the stability-plasticity dilemma \cite{grossberg1980}, that is the difficulty of a model to be both plastic enough to acquire new information and stable enough to preserve previously acquired knowledge.\\
Continual learning may have a large impact on a variety of real world applications: Computer vision \cite{lomonaco2017}, Natural Language Processing \cite{sun2020} and Robotics \cite{thrun1995} are examples of environments where the data is highly non stationary and may vary over time. A model able to learn continuously without forgetting would not need to be retrained from scratch every time a new experience is introduced (cumulative training). In fact, retraining is often the only viable alternative to continual learning in dynamic environments. However, retraining requires to store all the encountered data, which is often unfeasible under real world constraints. To address this problem, in this paper we focused on Replay strategies \cite{chaudhry2019a,aljundi2019b}, a family of CL techniques which leverages a buffer of patterns from previous experiences and uses it together with the current data to train the model.

We introduce a novel CL strategy called Distilled Replay to address the problem of building very small replay buffers with highly informative samples. Distilled Replay is based on the assumption that, if a replay pattern represents most of the features present in a dataset, it will be more effective against forgetting than a randomly sampled pattern from the dataset. Moreover, keeping a small buffer is useful to deploy continual learning solutions in real-world applications, since the memory footprint is drastically reduced with respect to cumulative approaches. Distilled Replay directly optimizes the memory consumption by keeping a buffer of only one pattern per class, while still retaining most of the original performance. The buffer is built using a distillation process based on Dataset Distillation \cite{Wang2018}, which allows to condensate an entire dataset into few informative patterns. Dataset Distillation removes the need to select replay patterns from the real dataset. Instead, it learns patterns which summarize the main characteristics of the dataset. Distilled Replay acts by combining our modified version of Dataset Distillation with replay strategies. It shows that even one pattern per class is sufficient to mitigate forgetting. In contrast, other replay strategies need larger memory buffers to match the performance of our approach.


\section{Continual Learning Scenario}
In this work, we consider continual learning on a sequence of $T$ experiences $E_1, \ldots, E_T$. Each experience $E_t$ is associated to a training set $D_t^{tr}$, a validation set $D_t^{vl}$ and a test set $D_t^{ts}$. 
A continual learning algorithm operating in the aforementioned scenario can be defined as follows \cite{lesort2020}:
\begin{equation}
\begin{aligned}
    &\forall D_t^{tr} \in S\ , \\
    &A_t^{CL}: \,\,<\boldsymbol{\theta}_{t-1}, D^{tr}_t, B_{t-1}, l_t>\ \rightarrow\ <\boldsymbol{\theta}_{t}, B_t>\ ,
\end{aligned} 
\end{equation}
where:
\begin{itemize}
    \item $\boldsymbol{\theta_t}$ are the model parameters at experience $t$, learned continually; 
    \item $B_t$ is an external buffer to store additional knowledge (like previously seen patterns);
    \item $l_t$ is an optional task label associated to each experience. The task label can be used to disentangle tasks and customize the model parameters (e.g. by using multi-headed models \cite{farquhar2019});
    \item $D^{tr}_t$ is the training set of examples. 
\end{itemize}

An algorithm respecting this formalization can be applied to different continual learning scenarios. In this paper, we used domain-incremental and class-incremental scenarios, identified in \cite{Ven2019}. 

In Domain Incremental Learning (D-IL) the classes to learn are all present from the first experience, but their generating distribution is subjected to a drift from one experience to the other.

In Class Incremental Learning (C-IL) scenarios each experience provides patterns coming from classes which are not present in the other experiences. 

We impose additional constraints to these scenarios, restricting the number of elements that can be stored from previous experiences to one per class and the number of epochs to one (single-pass). 

\section{Related Works}
\label{Section: rel_works}
The challenge of learning continuously has been addressed from different point of views \cite{lesort2020}. Regularization approaches try to influence the learning trajectory of a model in order to mitigate forgetting of previous knowledge \cite{chaudhry2018}. The regularization term is designed to increase the model stability across multiple experiences, for example by penalizing large changes in parameters deemed important for previous experiences \cite{kirkpatrick2017}.\\
Architectural strategies refer to a large set of techniques aimed at dynamically modifying the model structure . The addition of new components (e.g. layers \cite{rusu2016}) favors learning of new information, while forgetting can be mitigated by freezing previously added modules \cite{asghar2019} or by allocating separate components without interference \cite{rusu2016,cossu2020}. 
\paragraph{Dual memories strategies.}
This family of CL algorithms is loosely inspired by the Complementary Learning System theory (CLS) \cite{McClelland1995}.
This theory explains the memory consolidation process as the interplay between two structures: the hippocampus, responsible for the storage of recent episodic memories, and the neocortex, responsible for the storage of long-term knowledge and for the generalization to unseen events.

The idea of having an episodic memory (i.e. a buffer of previous patterns) which replays examples to a long-term storage (i.e. the model) is very popular in continual learning \cite{chaudhry2019a}. In fact, replay strategies are the most common representatives of dual memory strategies and very effective in class-incremental scenarios \cite{vandeven2020}. \\
Replay strategies are based on a sampling operation, which extracts a small buffer of patterns from each experience, and on a training algorithm which combines examples from the buffer with examples from the current experience. Sampling policies may vary from random selection to the use of heuristics to select patterns that are more likely to improve recall of previous experiences \cite{aljundi2019b}. Additionally, generative replay approaches \cite{shin2017} do not rely directly on patterns sampled from the dataset. Instead, they train a generative model to produce patterns similar to the ones seen at training time. Our approach share loose similarities with generative replay, since we do not replay patterns directly sampled from the original training set either. 
One important aspect of all replay approaches is the size of the replay buffer. Real-world applications may be constrained to the use of small buffers since they allow to scale to a large number of experiences. For this reason, one important objective of replay techniques is to minimize the buffer size \cite{chaudhry2019a}. However, relying on patterns sampled directly from the dataset or generated on-the-fly by a generative model may need many examples to mitigate forgetting. Our method, instead, leverages one of the simplest replay policies and manages to maintain one highly informative pattern per class in the buffer.



\section{Distilled Replay} \label{sec:approch}
Our proposed approach, called Distilled Replay, belongs to the family of dual memory strategies. Distilled Replay combines a simple replay policy with a small buffer composed of highly informative patterns. Instead of using raw replay samples, Distilled Replay learns the patterns to be replayed via buffer distillation, a process based on dataset distillation \cite{Wang2018}. In the remainder of this section, we describe the buffer distillation process and how to use it together with replay policies in continual learning.


\paragraph{Buffer distillation}
Replay strategies operating in the small buffer regime are forced to keep only few examples per class. These examples may not be representative of their own classes, thus reducing the effectiveness of the approach. Distilled Replay addresses this problem by implementing buffer distillation, a technique inspired by Dataset Distillation \cite{Wang2018}. Our buffer distillation compresses a dataset into a small number of highly informative, synthetic examples. Given a dataset $\boldsymbol{x} = \{x_i\}_{i=1}^N$, and an initialization $\boldsymbol{\theta_0}$, the goal is to learn a buffer of samples $\tilde{\boldsymbol{x}} = \{\tilde{x}_i\}_{i=1}^M$, initialized with samples from the dataset, and with $M\ll N$. Performing $S$ steps of SGD on the buffer $\boldsymbol{\tilde{x}}$ results in a model

\begin{equation}  
    \label{eq:distill_inner}
    \boldsymbol{\theta}_{S} = \boldsymbol{\theta}_{S-1} - \eta \nabla_{\boldsymbol{\theta}_{S - 1}} \ell(\tilde{\boldsymbol{x}}, \boldsymbol{\theta}_{S - 1}),
\end{equation}
that performs well on the original dataset $\boldsymbol{x}$. Buffer distillation achieves this result by solving the optimization problem:
\begin{equation} \label{eq:distill_outer}
    \tilde{\boldsymbol{x}}^* = \underset{\tilde{\boldsymbol{x}}}{arg\,min\,}\ \mathbb{E}_{\boldsymbol{\theta}_0 \sim p(\boldsymbol{\theta}_0)}\ \sum_{s=1}^{S} \ell (\boldsymbol{x} , \boldsymbol{\theta}_S),
\end{equation}
where $\ell (\boldsymbol{x} , \boldsymbol{\theta})$ is the loss for model $\boldsymbol{\theta}_S$ computed on $\boldsymbol{x}$. We can solve the optimization problem defined in Eq. \ref{eq:distill_outer} by stochastic gradient descent (the dependence of Eq. \ref{eq:distill_outer} on $\tilde{\boldsymbol{x}}$ is obtained by expanding $\boldsymbol{\theta}_S$ as in Eq. \ref{eq:distill_inner}). 
A model trained on synthetic samples $\tilde{\boldsymbol{x}}^*$ reduces the prediction loss on the original training set. Moreover, using a distribution of initializations $p(\boldsymbol{\theta}_0)$ makes the distilled samples independent from the specific model initialization values.

Algorithm \ref{Algorithm: dd_imp} shows the pseudocode for buffer distillation. We refer to the loop that updates the distilled images as outer loop and to the one that updates the model as inner loop. 

\begin{figure}[t]
    \centering
    \includegraphics[width=0.5\textwidth]{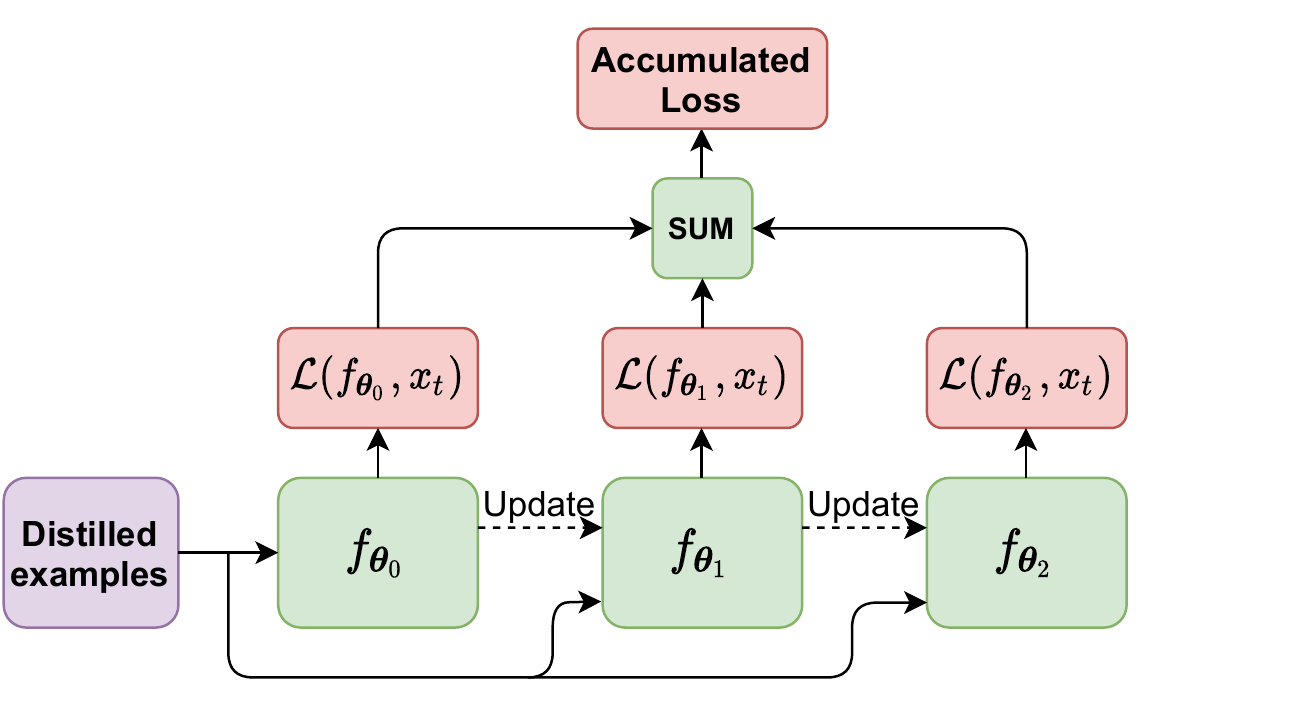}
    \caption{Schematic representation of Buffer Distillation update. The gradient is backpropagated to the distilled examples following the computation path defined by the solid arrows. The accumulated loss sums the loss computed on each step.}
    \label{Figure:bd}
\end{figure}

Our buffer distillation has some distinguishing characteristics with respect to the original Dataset Distillation \cite{Wang2018}.\\
While Dataset Distillation learns the learning rate $\eta$ used in the inner loop, we decided to fix it. In fact, Distilled Replay, during training, uses the same learning rate for past and current examples. If the examples were distilled to work with custom learning rates, they would lose their effectiveness when used with a different one. Therefore, during distillation the learning rate is the same used by the model during continual learning. \\
Another important difference with respect to Dataset Distillation is that our approach uses a different loss function. Instead of measuring the loss of the model on a minibatch of training data on the inner last step, our algorithm does that at each steps and backpropagate on their sum. As a result, buffer distillation is equivalent to backpropagating the gradient at each inner step and then updating the distilled images once the inner training is over. This process is graphically described by Figure \ref{Figure:bd}\\


We can summarize our buffer distillation in three steps:
\begin{enumerate}
    \item Do $S$ steps of gradient descent on the distilled images, obtaining $\boldsymbol{\theta}_1, \hdots, \boldsymbol{\theta}_S$.
    \item Evaluate each model on a minibatch of training data $x_t$ getting the loss $L_{tot} = \sum_{i=1}^{S} \ell(x_r, \boldsymbol{\theta}_i)$.
    \item Compute $\nabla_{\boldsymbol{\tilde{x}}} L_{tot}$ and update the distilled images.
\end{enumerate}

\begin{center}
\begin{algorithm}
\small
\caption{Buffer Distillation} \label{Algorithm: dd_imp}
\textbf{Input:} $p(\boldsymbol{\theta}_0)$: distribution of initial weights; $M$: number of distilled samples \\
\textbf{Input:} $\alpha$: step size; $R$: number of outer steps; $\eta$: learning rate \\
\textbf{Input:} $S$: number of inner inner steps. \\
\begin{algorithmic}[1]
\STATE Initialize $\tilde{\boldsymbol{x}} = \{\tilde{x}_i\}_{i = 1}^M$ 
\FORALL{outer steps $r = 1 \,\,\,\boldsymbol{to}\,\,\, R$}
    \STATE Get a minibatch of real training data $\boldsymbol{x}_r = \{x_{j}\}_{j=1}^n$
    \STATE Sample a batch of initial weights $\boldsymbol{\theta}_0^{(j)} \sim p(\boldsymbol{\theta}_0)$
    \FORALL{sampled $\boldsymbol{\theta}_0^{(j)}$}
    \FORALL{inner steps $s = 1 \,\,\,\boldsymbol{to}\,\,\, S$}
    \STATE Compute updated parameter with GD: $\boldsymbol{\theta}_s^{(j)} = \boldsymbol{\theta}_{s-1}^{(j)} - \eta \nabla_{\boldsymbol{\theta}_{s-1}^{(j)}} \ell (\tilde{\boldsymbol{x}}, \boldsymbol{\theta}_{s-1}^{(j)})$
    \STATE Evaluate the objective function on real training data: $\mathcal{L}^{(s, j)} = \ell(\boldsymbol{x}_r, \boldsymbol{\theta}_s^{(j)})$
    \ENDFOR
    \ENDFOR
    \STATE Update $\tilde{\boldsymbol{x}} \gets \tilde{\boldsymbol{x}} - \alpha \nabla_{\tilde{\boldsymbol{x}}}\sum_{s}\sum_{j} \mathcal{L}^{(s, j)}$
\ENDFOR
\end{algorithmic}
\textbf{Output:} distilled data $\tilde{\boldsymbol{x}}$
\end{algorithm}
\end{center}

\paragraph{Distilled Replay Training}

\begin{figure}[t]
    \centering
    \includegraphics[width=0.3\textwidth]{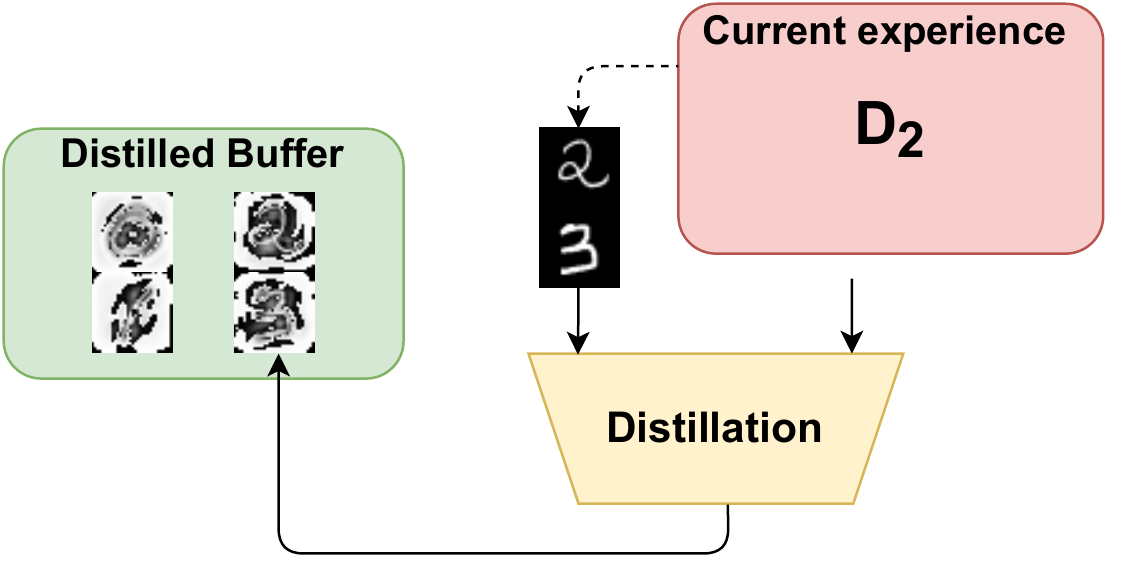}
    \caption{On experience $t$, a set of training examples is sampled from the current dataset, distilled and added to the buffer. The distilled samples will then be replayed alongside the next experience dataset.} 
    \label{Figure:dr}
\end{figure}

Distilled Replay combines a replay policy with the buffer learned by buffer distillation. Distilled Replay is designed to work with very small buffers, comprising as little as a single pattern per class.\\
During training, the replay policy builds each minibatch with elements sampled from the current dataset $D_t^{tr}$ and patterns from the buffer $B_t$. The combination of old and new data allows the model to learn the current experience while mitigating forgetting on past ones.\\
At the end of the $t$-th experience, we randomly sample a certain amount of elements per class from the current training set and use them to initialise the memory $\boldsymbol{m}_t$. We then apply buffer distillation to learn the synthetic memory $\tilde{\boldsymbol{m}}_t$. Finally, as shown in Figure \ref{Figure:dr}, we add the distillation result to the distilled buffer. Algorithm \ref{Algorithm: dr} shows the pseudocode for the entire training loop.

\section{Experiments}
We used the Average Accuracy \cite{chaudhry2019} as the main metric to monitor forgetting of previous knowledge. After training on the $t$-th experience, the average accuracy is evaluated by averaging on all experiences encountered so far: 
\begin{equation}
    \mathcal{A}_t = \frac{1}{t} \sum_{i=1}^t A(\boldsymbol{\theta}, D_i),
\end{equation}
where $\mathcal{A}(\boldsymbol{\theta}, D_i)$ is the accuracy on dataset $D_i$ from experience $E_i$ obtained with a model parameterized by $\boldsymbol{\theta}$.

We compared our approach against $5$ different continual learning strategies:
\begin{description}
    \item[Naive (LB)] trains the model continually without taking any measure to prevent forgetting. We use this strategy as a Lower Bound (LB) for continual learning performance.
    \item[Simple Replay (SR)] stores a buffer of examples randomly extracted from previous datasets and uses them to rehearse previous experiences. 
    \item[Cumulative (UB)] trains the model continually on the union of the datasets from the current and all the past experiences. Cumulative training keeps all the past data. Therefore, we used it as Upper Bound (UB) for the continual learning performance.
    \item[iCaRL \cite{rebuffi2017}] a dual-memory algorithm which combines knowledge distillation and nearest class mean classification.
    \item[Maximal Interfered Retrieval (MIR) \cite{aljundi2019b}] a dual-memory algorithm which selects the samples to be replayed based on how much their accuracy would drop after a training step on the current minibatch of data.
\end{description}
We measured the ability of the six strategies to prevent forgetting by using a single sample per class. We used a Multilayer Perceptron with one hidden layer of $500$ units in the D-IL scenario and a LeNet5 \cite{LeCun1998} in the C-IL scenario.
\begin{center}
\begin{algorithm}[t]
\caption{Distilled Replay Training} \label{Algorithm: dr}
\begin{algorithmic}[1]
\STATE $v \gets \text{list of datasets}$
\STATE $\emph{B}_0 \gets \text{list of memories}$
\FOR{$t = 1$ \TO $T$}
    \STATE $D^{tr}_t, D^{ts}_t \sim E_t$
    \STATE $v.\text{insert}(D_t^{ts})$
    \FOR{$q = 1$ \TO $Q$}
        \STATE $\text{mb}_q = \text{sample}(D_t^{tr})$ \hfill\COMMENT{Sample from dataset}
        \STATE $b_q = \emph{B}_t\ \cup \text{mb}_q $ \hfill\COMMENT{Create minibatch}
        \STATE $\boldsymbol{\theta}_q \gets \boldsymbol{\theta}_{q-1} - \nabla_{\boldsymbol{\theta}_{q-1}} \ell(b_q, \,\boldsymbol{\theta}_{q-1})$
    \ENDFOR
    \STATE $m_t \sim D_t^{tr}$
    \STATE $\tilde{m}_t \gets \text{buffer distillation}(m_t,\,D_t^{tr})$
    \STATE $\emph{B}_{t} \gets \emph{B}_{t-1} \cup \tilde{m}_t$
    \FORALL{$D^{ts}_i \,\,\,\boldsymbol{in}\,\,\, v$}
        \STATE $\text{test}(\boldsymbol{\theta}_{Q}, D_i^{ts})$
    \ENDFOR
\ENDFOR
\end{algorithmic}
\end{algorithm}
\end{center}

\begin{figure*}[t]
\centering
\footnotesize
\begin{subfigure}{.23\textwidth}
    \centering
    \begin{tabular}{l|ll||l||lll}
    \toprule
                     & \textbf{UB} & \textbf{LB} & \textbf{DR}   & \textbf{SR} & \textbf{iCaRL} & \textbf{MIR}\\ \midrule 
    $\boldsymbol{E_6}$  & .96 & .91 & \textbf{.93} & .91 & .92 & .91\\  
    $\boldsymbol{E_7}$  & .96 & .89 & \textbf{.92} & .90 & .91 & .90\\ 
    $\boldsymbol{E_8}$  & .96 & .87 & \textbf{.91} & .88 & .90 & .89\\ 
    $\boldsymbol{E_9}$  & .96 & .84 & \textbf{.91} & .88 & .89 & .85 \\ 
    $\boldsymbol{E_{10}}$ & .97 & .83 & \textbf{.90} & .88 & .89 & .83\\ 
    \bottomrule
    \end{tabular}
    \caption{Permuted MNIST}
    \label{Figure: a}
\end{subfigure}
\quad \quad \quad \quad \quad \quad \quad \quad
\quad \quad \quad \quad
\begin{subfigure}{.25\textwidth}
    \centering
    \begin{tabular}{l|ll||l||lll}
    \toprule
                     & \textbf{UB} & \textbf{LB} & \textbf{DR}   & \textbf{SR} & \textbf{iCaRL} & \textbf{MIR}\\ \midrule 
    $\boldsymbol{E_2}$  & .96 & .49 & \textbf{.93} & .88 & .91 & .92\\
    $\boldsymbol{E_3}$  & .94 & .33 & \textbf{.89} & .70 & .85 & .78\\
    $\boldsymbol{E_4}$  & .93 & .25 & \textbf{.87} & .66 & .81 & .68\\
    $\boldsymbol{E_5}$ & .91 & .20 & \textbf{.82} & .61 & .77 & .59\\ \bottomrule
    \end{tabular}
    \caption{Split MNIST}
    \label{Figure: b}
\end{subfigure}
\\
\begin{subfigure}{.24\textwidth}
    \centering
    \begin{tabular}{l|ll||l||lll}
    \toprule
                     & \textbf{UB} & \textbf{LB} & \textbf{DR}   & \textbf{SR} & \textbf{iCaRL} & \textbf{MIR}\\ \midrule 
    $\boldsymbol{E_2}$  & .93 & .50 & \textbf{.84} & .74 & .82 & .76\\
    $\boldsymbol{E_3}$  & .84 & .33 & \textbf{.67} & .55 & .66 & .55\\
    $\boldsymbol{E_4}$  & .76 & .30 & \textbf{.63} & .54 & .59 & .55\\
    $\boldsymbol{E_5}$  & .78 & .19 & \textbf{.63} & .48 & .60 & .45\\
    \bottomrule
    \end{tabular}
    \caption{Split Fashion MNIST}
    \label{Figure: c}
\end{subfigure}
\quad \quad \quad \quad \quad \quad \quad \quad
\quad \quad \quad \quad
\begin{subfigure}{.23\textwidth}
    \centering
    \begin{tabular}{l|ll||l||lll}
    \toprule
                     & \textbf{UB} & \textbf{LB} & \textbf{DR}   & \textbf{SR} & \textbf{iCaRL} & \textbf{MIR}\\ \midrule 
    $\boldsymbol{E_2}$  & .56 & .28 & \textbf{.52} & .43 & .41 & .49\\
    $\boldsymbol{E_3}$  & .43 & .21 & \textbf{.34} & .29 & .29 & .32\\
    $\boldsymbol{E_4}$  & .38 & .18 & \textbf{.28} & .21 & .23 & .18\\
    $\boldsymbol{E_5}$  & .35 & .14 & \textbf{.24} & .19 & 21 & .19\\ \bottomrule
    \end{tabular}
    \caption{Split CIFAR-10}
    \label{Figure: d}
\end{subfigure}
\captionof{table}{Average accuracies of the six tested methods on the four benchmarks. The leftmost column of each table reports the experience up to which the accuracy is averaged. For Permuted MNIST, we report the average accuracy starting from the $6$-th experience. In fact, the last experiences better represents the overall performance, since the accuracy is averaged over all experiences seen so far.}
\label{Figure: all-methods}
\end{figure*}

We experimented with four popular continual learning benchmarks for image classification: Permuted MNIST \cite{Goodfellow2015}, Split MNIST \cite{Zenke2017b}, Split Fashion MNIST \cite{Xiao2017} and Split CIFAR10 \cite{Paz2017}.\\
Permuted MNIST is a Domain-incremental scenario in which each experience is constructed by applying a fixed random permutation to all the MNIST images. The permutation only changes at the end of each experience. We used $10$ experiences in total.
The other benchmarks are class-incremental benchmarks in which each experience is composed by examples from two classes. In this setup, the number of classes increases every time a new experience is introduced. Therefore, the number of experiences is $5$ for each benchmark.

\subsection{Results}
Table \ref{Figure: all-methods} reports the Average Accuracy after training on each experience for all the evaluated strategies. Distilled Replay consistently outperforms the other methods, often by a large margin. In particular, neither iCaRL nor MIR are able to surpass Distilled Replay in the challenging continual learning scenario used in our experiments (single-epoch, replay buffers with one pattern per class).
On Permuted MNIST, after the last experience, the accuracies of the compared methods drop between $83\%$ and $89\%$. Our Distilled Replay is able to reach around $90\%$ accuracy at the end of the last experience. 

\begin{figure}
    \centering
    \includegraphics[width=.4\textwidth]{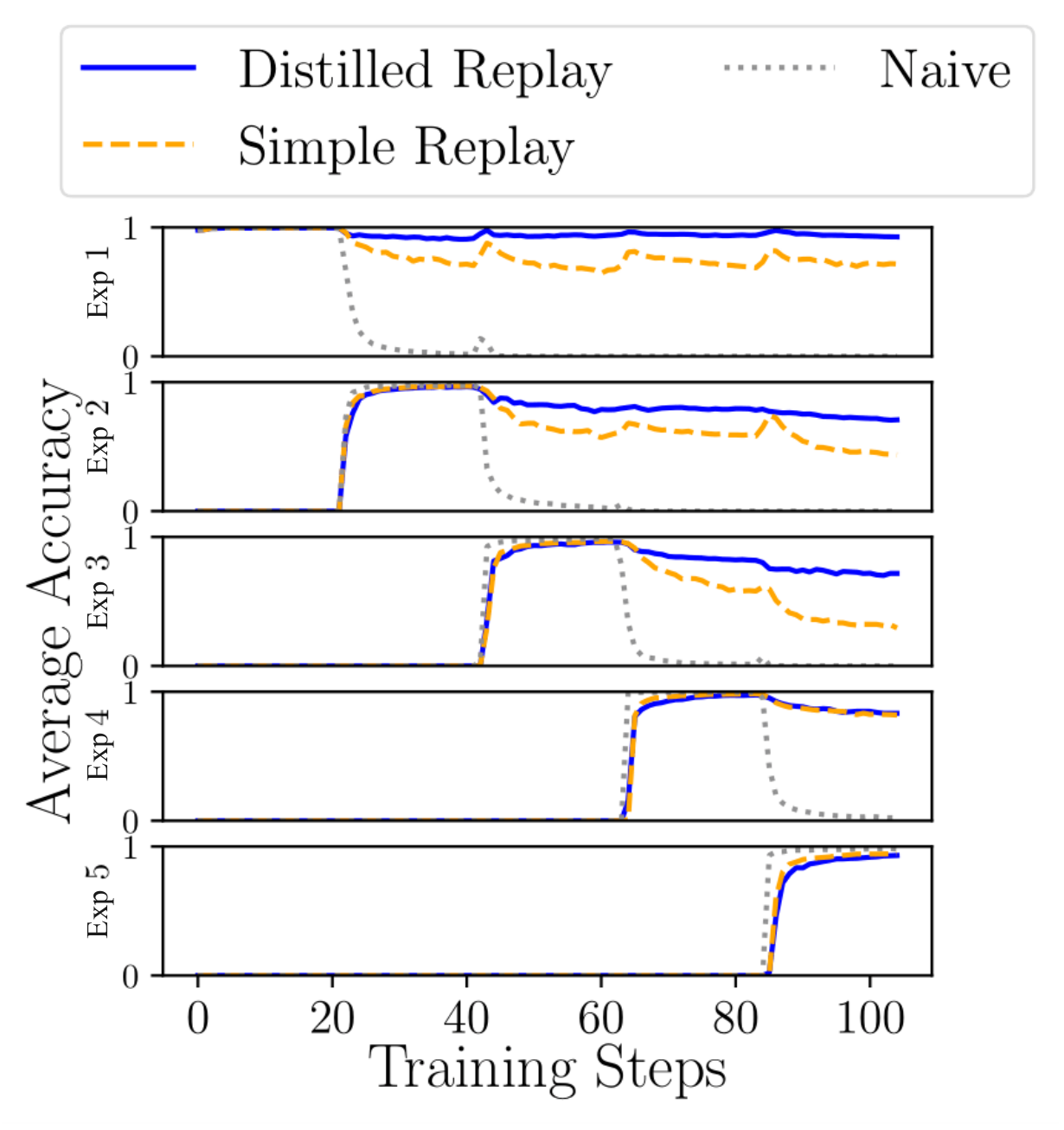}
    \caption{Accuracies on each S-MNIST experience. Without replaying any data, the performance on previous experiences drop. By using distilled replay, we manage to get higher performance compared to standard replay.}
    \label{Figure: tiled}
\end{figure}
\begin{figure}
    \centering
    \subfloat{{\includegraphics[width=3 cm]{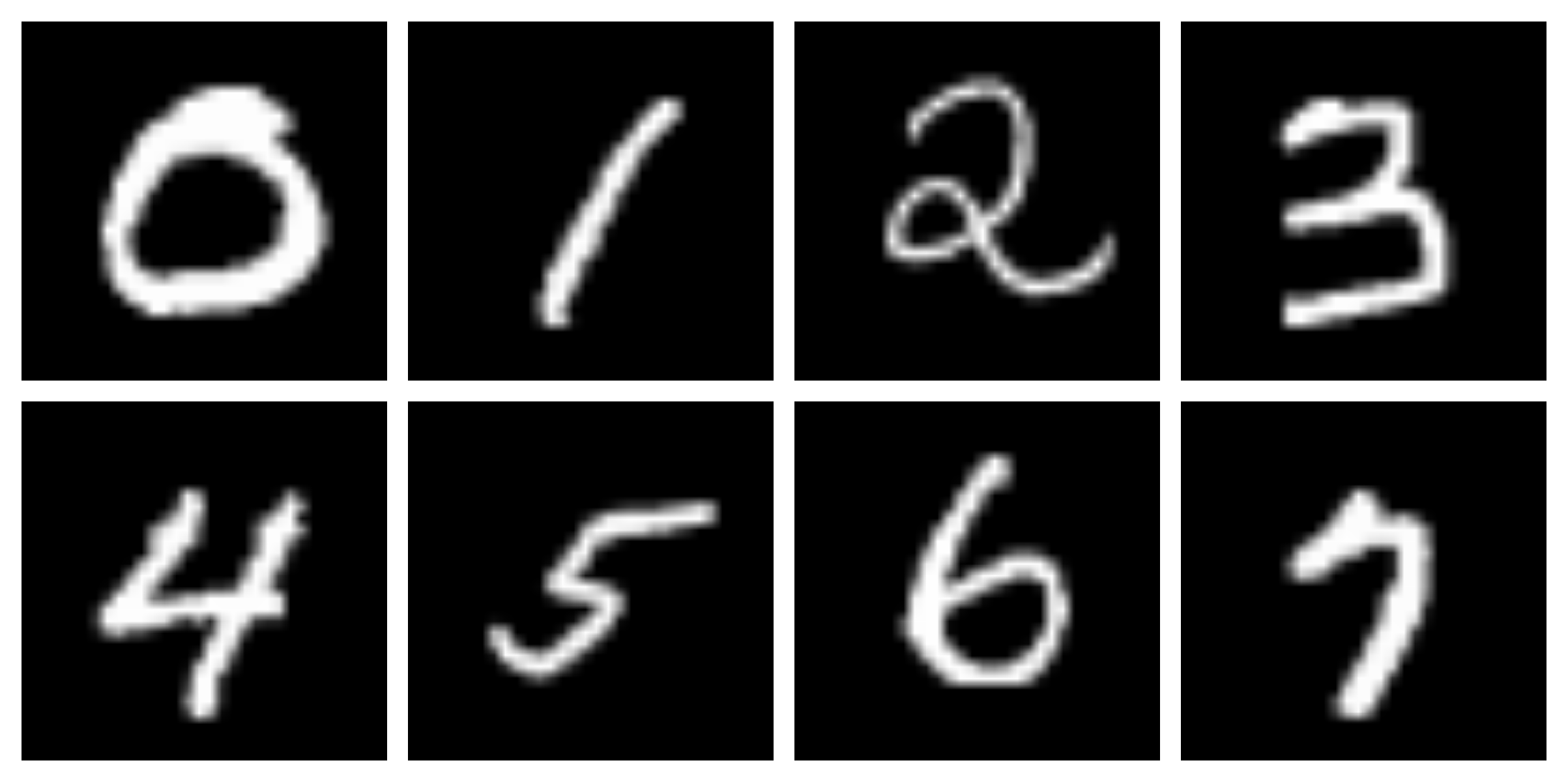} }}%
    \qquad
    \subfloat{{\includegraphics[width=3 cm]{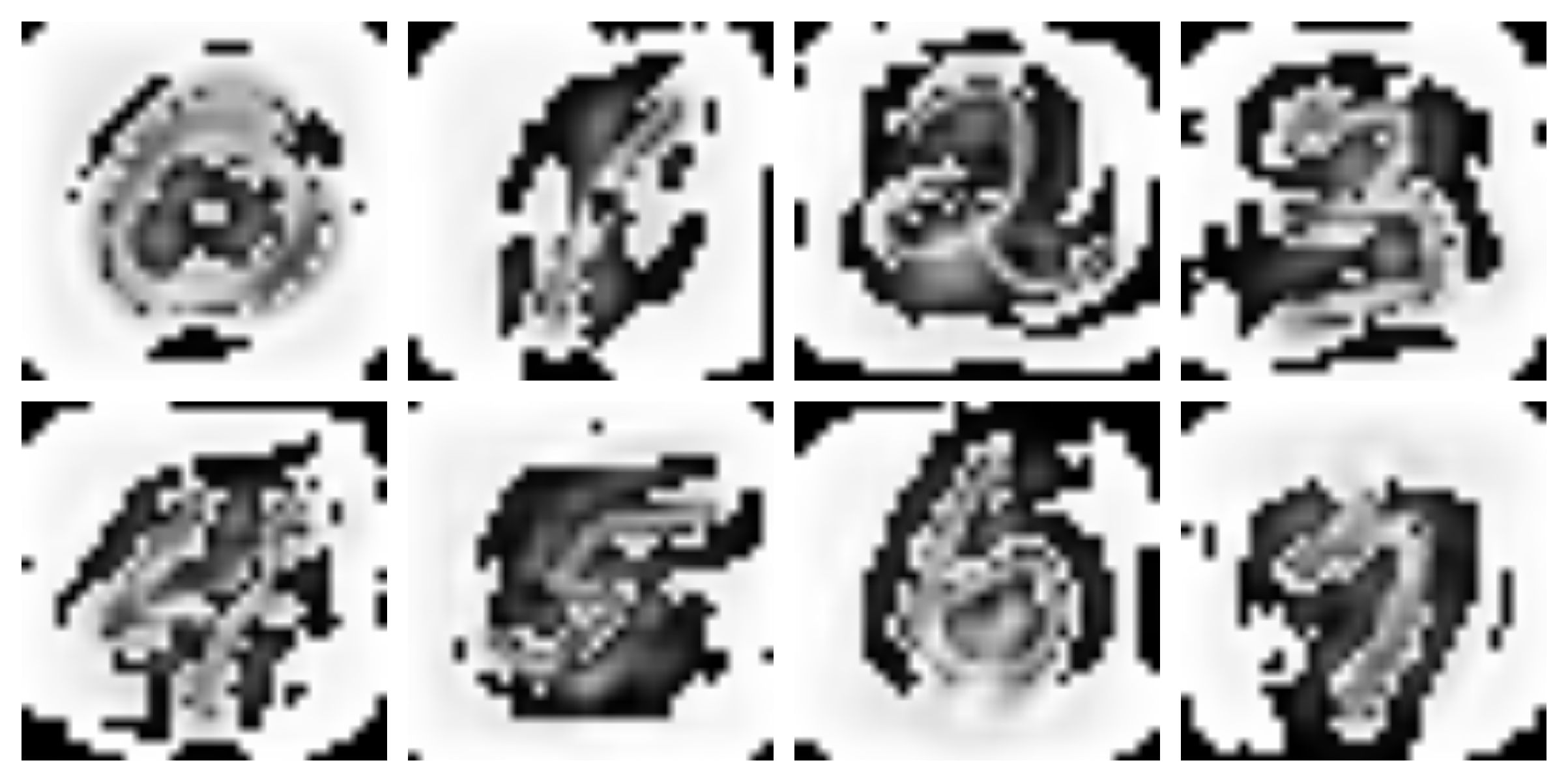} }}%
    \caption{Replay memory content of Simple Replay (left) and Distilled Replay (right). Distilled samples highlight the most representative features of the input patterns.}%
    \label{fig:exp3}%
\end{figure}

In C-IL benchmarks, Distilled Replay outperforms the other strategies by a larger margins than in the D-IL scenario. Table \ref{Figure: b} shows the performance on Split MNIST. iCaRL and MIR obtained $77\%$ and $59\%$ accuracy respectively, while Distilled Replay achieves an accuracy of $82\%$. 
Figure \ref{fig:exp3} shows the patterns in the replay buffer of Simple Replay and Distilled Replay. The ones used by standard replay are simply patterns taken from the training datasets. The patterns in the buffer of Distilled Replay shows a white background and a thicker digit contour. The performance of Distilled Replay and Simple Replay differs not only in accuracy values but also in their trajectories: Simple Replay accuracy degrades faster as the training progresses. To highlight this phenomenon, Figure \ref{Figure: tiled} reports the accuracies on each experience throughout learning. We can see how dataset distillation maintains a higher accuracy on past experiences. This results in a more stable average accuracy.\\
More challenging C-IL benchmarks such as Split Fashion MNIST (Table \ref{Figure: c}) and Split CIFAR-10 (Table \ref{Figure: d}) show similar differences between Distilled replay performances and the ones of the compared strategies. Distilled Replay outperforms the other methods, but the absolute performance of all six strategies is lower than on Split MNIST.\\
From Table \ref{Figure: d}, we can observe that on Split CIFAR-10 there is a large drop in performance compared to the previous benchmarks. This is consistent for all the evaluated strategies. In Section \ref{sec:discussion} we highlight some of the issues that may explain the reduced performance of Distilled Replay on challenging data consisting of complex patterns.

\subsection{Ablation Study}
Our buffer distillation process introduces significant differences as compared to the original Dataset Distillation technique \cite{Wang2018} (see Section \ref{sec:approch}). Therefore, we ran experiments on Split MNIST to validate the importance of these modifications in a continual learning scenario. 
The hyperparameters have been selected as follows. We kept the same number of inner and outer steps of distillation in both algorithms, so that the computation time is approximately equal. Instead, the learning rate of the outer update was selected by validating on different values (i.e. $0.05$, $0.1$, $0.5$). In particular, we found  that increasing the outer learning rates from $0.1$ to $0.5$ led to a better performance in Dataset Distillation.

\begin{figure}[t]
    \centering
    \includegraphics[width=0.3\textwidth]{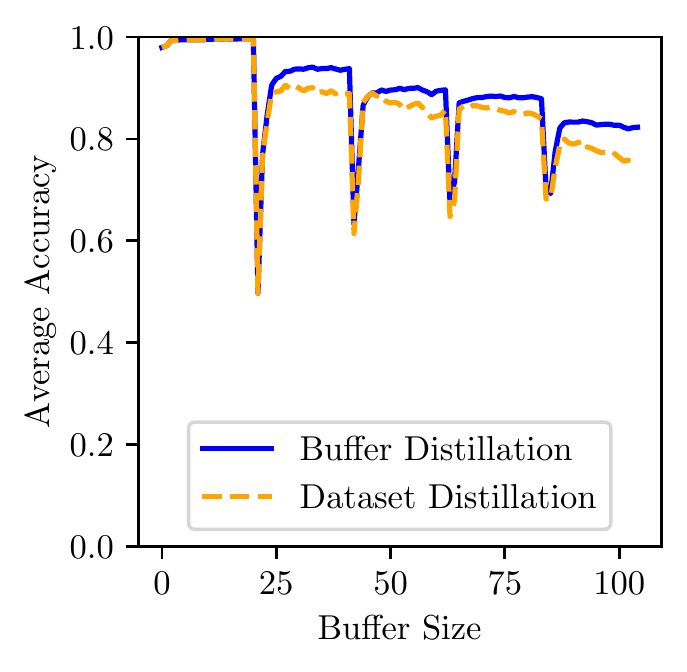}
    \caption{Comparison of two versions of the distilled replay algorithm on Split MNIST. One version is based on the original Dataset Distillation algorithm \protect\cite{Wang2018}, while the other uses our buffer distillation.}
    \label{Figure: no_imp}
\end{figure}

Figure \ref{Figure: no_imp} shows the learning curves of the Dataset Distillation and Buffer Distillation techniques. As soon as the model starts learning on the second experience, our distillation process outperforms Dataset Distillation, supporting the observation that the distilled samples have higher quality.

\subsection{Computational Times}
The Buffer Distillation process scales linearly with the number of inner steps. In a continual learning setting, we continually adapt the model and, consequently, the buffer. This requires the model to be able to learn from multiple passes over data. Therefore, we experimented with large values for the inner steps. However, the high number of steps in the inner loop contributed to increase the computational cost of the distillation.
\begin{figure}[t]
\centering
\begin{subfigure}{.22\textwidth}
    \centering
    \includegraphics[width=0.9\textwidth]{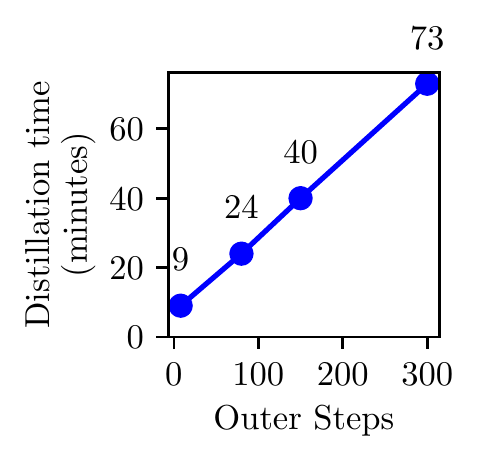}
\end{subfigure}
\begin{subfigure}{.22\textwidth}
    \centering
    \includegraphics[width=0.7\textwidth]{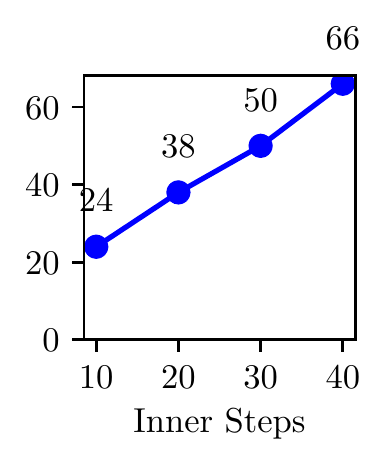}
\end{subfigure}%
\caption{Average time for a single buffer distillation on Split CIFAR10. On the left, we show the time as a function of the number of outer steps (inner steps fixed to $10$). On the right, we show the time as a function of the inner steps (outer steps fixed to $80$).}
\label{Figure: times}
\end{figure}
Notice that, differently from other popular continual learning strategies (e.g. GEM \cite{Paz2017}), whose computation mainly occurs during continual training, the distillation process is independent of the training on the current experience. Therefore, it is possible to perform Buffer Distillation in parallel to training as soon as a new experience arrives.
Figure \ref{Figure: times} reports the average time of the distillation process executed on a GPU Nvidia V100. The data comes from the distillation of Split CIFAR-10 experiences. In our experiments, the best configurations in terms of final average accuracy used $80$ outer steps and $20$ inner steps, with an average time of $38$ minutes for each buffer distillation. While this is not a prohibitive amount of computational time, it has to be multiplied by the number of experiences in the stream (except for the last one which is not distilled), making the Distilled Replay a relatively expensive strategy in terms of computational times.


\section{Discussion} \label{sec:discussion}
The main objective of our experimental analysis was to test whether replay strategies with small buffers of one pattern per class were able to mitigate forgetting.
\footnote{The code along with the configuration files needed to reproduce
our results are available at \href{https://github.com/andrew-r96/
DistilledReplay}{https://github.com/andrew-r96/
DistilledReplay}}
The results show that, in this small buffer regime, the use of real patterns sampled from the dataset may not be sufficient to recover performance of previous experiences. Instead, we show that highly informative samples generated by Buffer Distillation allow to mitigate forgetting. 
Building on our results, we can also identify some additional insights and issues of Distilled Replay worth exploring in further developments.
\paragraph{Independence of distillation processes.}
Since the buffer distillation process is applied separately for each experience, the synthetic samples are optimized without taking into account previous (and future) experiences. This makes the distillation process easier but it also brings possible downsides. For example, distilling samples of similar classes belonging to different experiences may introduce ambiguous features and increase the forgetting on such classes. Since we keep a single example per class, similar samples of different classes would negatively impact the final accuracy.
Datasets such as Fashion MNIST, containing a high number of classes similar to each other (e.g. t-shirt, shirt, coat, dress, pullover), may be affected by this problem.

\paragraph{Distilled Replay with complex architectures.} 
The results we showed for C-IL benchmarks use a LeNet5 architecture \cite{LeCun1998}. To improve the results on S-CIFAR-10 we did some preliminary experiments using a ResNet architecture \cite{He2016} together with Distilled Replay. However, we were not able to distill useful samples. The results (not shown in the paper) suggest that the buffer distillation process struggles with more complex architectures.
We hypothesize that the optimization of the distilled examples is too challenging for gradient descent on sufficiently complex models. In fact, the distillation objective requires the backpropagation of the gradient through multiple gradient descent steps. For sufficiently complex architectures, this would result in a large computational graph which may suffer from vanishing or exploding gradients issues \cite{Hochreiter1991}.

\paragraph{Robustness to continual training.}
Buffer Distillation is able to learn robust samples which better mitigate forgetting than the ones generated by the original Dataset Distillation. 
This is mainly due to the fact that buffer distillation optimizes the loss for each point of the learning trajectory. Therefore, the outer updates takes into consideration to what extent each distilled image influences the learning trajectory of the model.
As a result, buffer distillation produces patterns which are robust to small parameter changes. This is important in a continual learning setting, where the model must continually adapt to novel experiences.

\section{Conclusion and Future Work}

Replay based methods are among the most effective continual learning strategies. In this work, we introduced a novel replay strategy called Distilled Replay which combines replay with Buffer Distillation, a process that generates a small buffer of highly informative samples. In particular, we studied whether keeping in the buffer a single pattern per class is sufficient to recover most of the original performance. When compared to other replay strategies like iCaRL and MIR, Distilled Replay shows superior results. The ability of Buffer Distillation to learn highly informative patterns is crucial to boost the performance of replay with small buffers. \\
By leveraging recent works about novel dataset condensation mechanisms \cite{Zhao2021a,Zhao2021b}, it would be possible to improve the computational efficiency of Distilled Replay. 
Future works could also study the performance of Distilled Replay in very constrained settings where it is required to store less than one pattern per class, for example by iteratively applying distillation on the buffer itself.
Ultimately, we hope that our work will foster the study of replay strategies in the small buffer regime, where it is only possible to store few patterns. This would help in the development of more efficient and sustainable continual learning systems, able to operate in the real-world constrained settings.






{\small
\bibliographystyle{named}
\bibliography{CL,bibliography}

\begin{thebibliography}{}

\bibitem[\protect\citeauthoryear{Aljundi \bgroup \em et al.\egroup
  }{2019}]{aljundi2019b}
Rahaf Aljundi, Eugene Belilovsky, Tinne Tuytelaars, Laurent Charlin, Massimo
  Caccia, Min Lin, and Lucas {Page-Caccia}.
\newblock Online {{Continual Learning}} with {{Maximal Interfered Retrieval}}.
\newblock In {\em Advances in {{Neural Information Processing Systems}} 32},
  pages 11849--11860, 2019.

\bibitem[\protect\citeauthoryear{Asghar \bgroup \em et al.\egroup
  }{2019}]{asghar2019}
Nabiha Asghar, Lili Mou, Kira~A Selby, Kevin~D Pantasdo, Pascal Poupart, and
  Xin Jiang.
\newblock Progressive {{Memory Banks}} for {{Incremental Domain Adaptation}}.
\newblock In {\em International {{Conference}} on {{Learning
  Representations}}}, 2019.

\bibitem[\protect\citeauthoryear{Chaudhry \bgroup \em et al.\egroup
  }{2018}]{chaudhry2018}
Arslan Chaudhry, Puneet~K. Dokania, Thalaiyasingam Ajanthan, and Philip H.~S.
  Torr.
\newblock Riemannian {{Walk}} for {{Incremental Learning}}: {{Understanding
  Forgetting}} and {{Intransigence}}.
\newblock In {\em Proceedings of the {{European Conference}} on {{Computer
  Vision}} ({{ECCV}})}, pages 532--547, 2018.

\bibitem[\protect\citeauthoryear{Chaudhry \bgroup \em et al.\egroup
  }{2019a}]{chaudhry2019}
Arslan Chaudhry, Marc'Aurelio Ranzato, Marcus Rohrbach, and Mohamed Elhoseiny.
\newblock Efficient {{Lifelong Learning}} with {{A}}-{{GEM}}.
\newblock In {\em {{ICLR}}}, 2019.

\bibitem[\protect\citeauthoryear{Chaudhry \bgroup \em et al.\egroup
  }{2019b}]{chaudhry2019a}
Arslan Chaudhry, Marcus Rohrbach, Mohamed Elhoseiny, Thalaiyasingam Ajanthan,
  Puneet~K Dokania, Philip H~S Torr, and Marc'Aurelio Ranzato.
\newblock On {{Tiny Episodic Memories}} in {{Continual Learning}}.
\newblock {\em arXiv}, 2019.

\bibitem[\protect\citeauthoryear{Cossu \bgroup \em et al.\egroup
  }{2020}]{cossu2020}
Andrea Cossu, Antonio Carta, and Davide Bacciu.
\newblock Continual {{Learning}} with {{Gated Incremental Memories}} for
  sequential data processing.
\newblock In {\em Proceedings of the 2020 {{International Joint Conference}} on
  {{Neural Networks}} ({{IJCNN}} 2020)}, 2020.

\bibitem[\protect\citeauthoryear{Farquhar and Gal}{2019}]{farquhar2019}
Sebastian Farquhar and Yarin Gal.
\newblock Towards {{Robust Evaluations}} of {{Continual Learning}}.
\newblock In {\em Privacy in {{Machine Learning}} and {{Artificial
  Intelligence}} Workshop, {{ICML}}}, 2019.

\bibitem[\protect\citeauthoryear{French}{1999}]{french1999}
Robert~M. French.
\newblock Catastrophic forgetting in connectionist networks.
\newblock {\em Trends in Cognitive Sciences}, 3:128--135, 4 1999.

\bibitem[\protect\citeauthoryear{Goodfellow \bgroup \em et al.\egroup
  }{2015}]{Goodfellow2015}
Ian~J. Goodfellow, Mehdi Mirza, Da~Xiao, Aaron Courville, and Yoshua Bengio.
\newblock An empirical investigation of catastrophic forgetting in
  gradient-based neural networks, 2015.

\bibitem[\protect\citeauthoryear{Grossberg}{1980}]{grossberg1980}
Stephen Grossberg.
\newblock How does a brain build a cognitive code?
\newblock {\em Psychological Review}, 87(1):1--51, 1980.

\bibitem[\protect\citeauthoryear{{He} \bgroup \em et al.\egroup
  }{2016}]{He2016}
K.~{He}, X.~{Zhang}, S.~{Ren}, and J.~{Sun}.
\newblock Deep residual learning for image recognition.
\newblock In {\em 2016 IEEE Conference on Computer Vision and Pattern
  Recognition (CVPR)}, pages 770--778, 2016.

\bibitem[\protect\citeauthoryear{Hochreiter}{1991}]{Hochreiter1991}
Sepp Hochreiter.
\newblock Untersuchungen zu dynamischen neuronalen netzen, 1991.

\bibitem[\protect\citeauthoryear{Kirkpatrick \bgroup \em et al.\egroup
  }{2017}]{kirkpatrick2017}
James Kirkpatrick, Razvan Pascanu, Neil Rabinowitz, Joel Veness, Guillaume
  Desjardins, Andrei~A Rusu, Kieran Milan, John Quan, Tiago Ramalho, Agnieszka
  {Grabska-Barwinska}, Demis Hassabis, Claudia Clopath, Dharshan Kumaran, and
  Raia Hadsell.
\newblock Overcoming catastrophic forgetting in neural networks.
\newblock {\em PNAS}, 114(13):3521--3526, 2017.

\bibitem[\protect\citeauthoryear{Krizhevsky \bgroup \em et al.\egroup
  }{2012}]{Krizhevsky2012}
Alex Krizhevsky, Ilya Sutskever, and Geoffrey~E Hinton.
\newblock Imagenet classification with deep convolutional neural networks.
\newblock In F.~Pereira, C.~J.~C. Burges, L.~Bottou, and K.~Q. Weinberger,
  editors, {\em Advances in Neural Information Processing Systems}, volume~25.
  Curran Associates, Inc., 2012.

\bibitem[\protect\citeauthoryear{LeCun \bgroup \em et al.\egroup
  }{1998}]{LeCun1998}
Yann LeCun, Léon Bottou, Yoshua Bengio, and Patrick Haffner.
\newblock Gradient-based learning applied to document recognition.
\newblock {\em Proceedings of the IEEE}, 86:2278--2323, 1998.

\bibitem[\protect\citeauthoryear{Lesort \bgroup \em et al.\egroup
  }{2020}]{lesort2020}
Timothée Lesort, Vincenzo Lomonaco, Andrei Stoian, Davide Maltoni, David
  Filliat, and Natalia Díaz-Rodríguez.
\newblock Continual learning for robotics: Definition, framework, learning
  strategies, opportunities and challenges.
\newblock {\em Information Fusion}, 58:52--68, 6 2020.

\bibitem[\protect\citeauthoryear{Lomonaco and Maltoni}{2017}]{lomonaco2017}
Vincenzo Lomonaco and Davide Maltoni.
\newblock {{CORe50}}: A {{New Dataset}} and {{Benchmark}} for {{Continuous
  Object Recognition}}.
\newblock In {\em Proceedings of the 1st {{Annual Conference}} on {{Robot
  Learning}}}, volume~78, pages 17--26, 2017.

\bibitem[\protect\citeauthoryear{Lopez-Paz and Ranzato}{2017}]{Paz2017}
David Lopez-Paz and Marc'Aurelio Ranzato.
\newblock Gradient episodic memory for continual learning.
\newblock {\em Advances in Neural Information Processing Systems},
  2017-December:6468--6477, 6 2017.

\bibitem[\protect\citeauthoryear{McClelland \bgroup \em et al.\egroup
  }{1995}]{McClelland1995}
James~L. McClelland, Bruce~L. McNaughton, and Randall~C. O'Reilly.
\newblock Why there are complementary learning systems in the hippocampus and
  neocortex: Insights from the successes and failures of connectionist models
  of learning and memory.
\newblock {\em Psychological Review}, 102:419--457, 1995.

\bibitem[\protect\citeauthoryear{Rebuffi \bgroup \em et al.\egroup
  }{2017}]{rebuffi2017}
Sylvestre-Alvise Rebuffi, Alexander Kolesnikov, Georg Sperl, and Christoph~H
  Lampert.
\newblock {{iCaRL}}: {{Incremental Classifier}} and {{Representation
  Learning}}.
\newblock In {\em The {{IEEE Conference}} on {{Computer Vision}} and {{Pattern
  Recognition}} ({{CVPR}})}, 2017.

\bibitem[\protect\citeauthoryear{Rusu \bgroup \em et al.\egroup
  }{2016}]{rusu2016}
Andrei~A Rusu, Neil~C Rabinowitz, Guillaume Desjardins, Hubert Soyer, James
  Kirkpatrick, Koray Kavukcuoglu, Razvan Pascanu, and Raia Hadsell.
\newblock Progressive {{Neural Networks}}.
\newblock {\em arXiv}, 2016.

\bibitem[\protect\citeauthoryear{Shin \bgroup \em et al.\egroup
  }{2017}]{shin2017}
Hanul Shin, Jung~Kwon Lee, Jaehong Kim, and Jiwon Kim.
\newblock Continual {{Learning}} with {{Deep Generative Replay}}.
\newblock In I~Guyon, U~V Luxburg, S~Bengio, H~Wallach, R~Fergus,
  S~Vishwanathan, and R~Garnett, editors, {\em Advances in {{Neural Information
  Processing Systems}} 30}, pages 2990--2999. {Curran Associates, Inc.}, 2017.

\bibitem[\protect\citeauthoryear{Sun \bgroup \em et al.\egroup
  }{2020}]{sun2020}
Fan-Keng Sun, Cheng-Hao Ho, and Hung-Yi Lee.
\newblock {{LAMOL}}: {{LAnguage MOdeling}} for {{Lifelong Language Learning}}.
\newblock In {\em {{ICLR}}}, 2020.

\bibitem[\protect\citeauthoryear{Thrun}{1995}]{thrun1995}
Sebastian Thrun.
\newblock A {{Lifelong Learning Perspective}} for {{Mobile Robot Control}}.
\newblock In Volker Graefe, editor, {\em Intelligent {{Robots}} and
  {{Systems}}}, pages 201--214. {Elsevier Science B.V.}, {Amsterdam}, 1995.

\bibitem[\protect\citeauthoryear{van~de Ven and Tolias}{2019}]{Ven2019}
Gido~M. van~de Ven and Andreas~S. Tolias.
\newblock Three scenarios for continual learning.
\newblock {\em arXiv}, 4 2019.

\bibitem[\protect\citeauthoryear{van~de Ven \bgroup \em et al.\egroup
  }{2020}]{vandeven2020}
Gido~M van~de Ven, Hava~T Siegelmann, and Andreas~S Tolias.
\newblock Brain-inspired replay for continual learning with artificial neural
  networks.
\newblock {\em Nature Communications}, 11:4069, 2020.

\bibitem[\protect\citeauthoryear{Wang \bgroup \em et al.\egroup
  }{2018}]{Wang2018}
Tongzhou Wang, Jun-Yan Zhu, Antonio Torralba, and Alexei~A. Efros.
\newblock Dataset distillation.
\newblock {\em arXiv}, 11 2018.

\bibitem[\protect\citeauthoryear{Xiao \bgroup \em et al.\egroup
  }{2017}]{Xiao2017}
Han Xiao, Kashif Rasul, and Roland Vollgraf.
\newblock Fashion-mnist: a novel image dataset for benchmarking machine
  learning algorithms.
\newblock {\em arXiv}, 8 2017.

\bibitem[\protect\citeauthoryear{Zenke \bgroup \em et al.\egroup
  }{2017}]{Zenke2017b}
Friedemann Zenke, Ben Poole, and Surya Ganguli.
\newblock Continual {{Learning Through Synaptic Intelligence}}.
\newblock In {\em International {{Conference}} on {{Machine Learning}}}, pages
  3987--3995, 2017.

\bibitem[\protect\citeauthoryear{Zhao and Bilen}{2021}]{Zhao2021b}
Bo~Zhao and Hakan Bilen.
\newblock Dataset condensation with differentiable siamese augmentation, 2021.

\bibitem[\protect\citeauthoryear{Zhao \bgroup \em et al.\egroup
  }{2021}]{Zhao2021a}
Bo~Zhao, Konda~Reddy Mopuri, and Hakan Bilen.
\newblock Dataset condensation with gradient matching, 2021.

\end{thebibliography}
}

\end{document}